\documentclass{sigkddExp}

\begin{document}
%

\title{Exploring Large Language Models for Feature Selection: \\A Data-centric Perspective}
%

\numberofauthors{3}
%


\author{
\alignauthor Dawei Li\thanks{Equal Constributions} \\
       \affaddr{Arizona State University}\\
       \affaddr{Tempe, AZ, USA}\\
       \email{daweili5@asu.edu}
\alignauthor Zhen Tan$^{*}$ \\
       \affaddr{Arizona State University}\\
       \affaddr{Tempe, AZ, USA}\\
       \email{ztan36@asu.edu}
\alignauthor Huan Liu \\
       \affaddr{Arizona State University}\\
       \affaddr{Tempe, AZ, USA}\\
       \email{huanliu@asu.edu}
}
\maketitle
\begin{abstract}
The rapid advancement of Large Language Models (LLMs) has significantly influenced various domains, leveraging their exceptional few-shot and zero-shot learning capabilities. 
In this work, we aim to explore and understand the LLMs-based feature selection methods from a data-centric perspective.
We begin by categorizing existing feature selection methods with LLMs into two groups: data-driven feature selection which requires numerical values of samples to do statistical inference and text-based feature selection which utilizes prior knowledge of LLMs to do semantical associations using descriptive context.
We conduct experiments in both classification and regression tasks with LLMs in various sizes (e.g., GPT-4, ChatGPT and LLaMA-2).
Our findings emphasize the effectiveness and robustness of text-based feature selection methods and showcase their potentials using a real-world medical application. We also discuss the challenges and future opportunities in employing LLMs for feature selection, offering insights for further research and development in this emerging field.

\end{abstract}

\section{Introduction}
\label{Introduction}
\vspace{0.35cm}

Recent years have witnessed the remarkable development of Large Language Models (LLMs)~\cite{achiam2023gpt,brown2020language,tan2024large,touvron2023llama} across various domains and areas~\cite{liang2022holistic,chang2023survey,li2024facial,beigi2024lrq}.
By leveraging extensive training corpora and well-designed prompting strategies, LLMs demonstrate impressive few-shot and zero-shot capabilities in diverse tasks such as question answering~\cite{wei2022chain,wangself,tong2024can}, information extraction~\cite{wadhwa2023revisiting} and knowledge discovery~\cite{pan2024unifying,wang2024large,wang2023noise}. 
The tuning-free nature also makes in-context learning (ICL) in LLMs achieve a great balance between efficiency and effectiveness~\cite{tan2024tuning}.

Feature selection~\cite{dash1997feature,li2017feature} is a critical data serving step that ensures relevant and high-quality data for downstream machine learning and data mining applications.
While existing data-driven selection methods have achieved great success in scenarios with abundant data and metadata, there is an increasing demand for efficient feature selection methods with few or even zero samples for various reasons~\cite{zhang2019correlated}.
This need is particularly pronounced in sensitive applications such as predicting survival times for cancer patients~\cite{tomczak2015review, wissel2022survboard}, where privacy concerns may prevent hospitals and patients from sharing their data, posing difficulties in the feature selection and engineering process.
To address this challenge, recent studies~\cite{jeong2024llm, hanlarge} have explored leveraging the few-shot capability in LLMs to perform feature selection in low-resource settings and got promising results. 

\begin{figure}[!t]
    \centering
    \includegraphics[width=8cm]{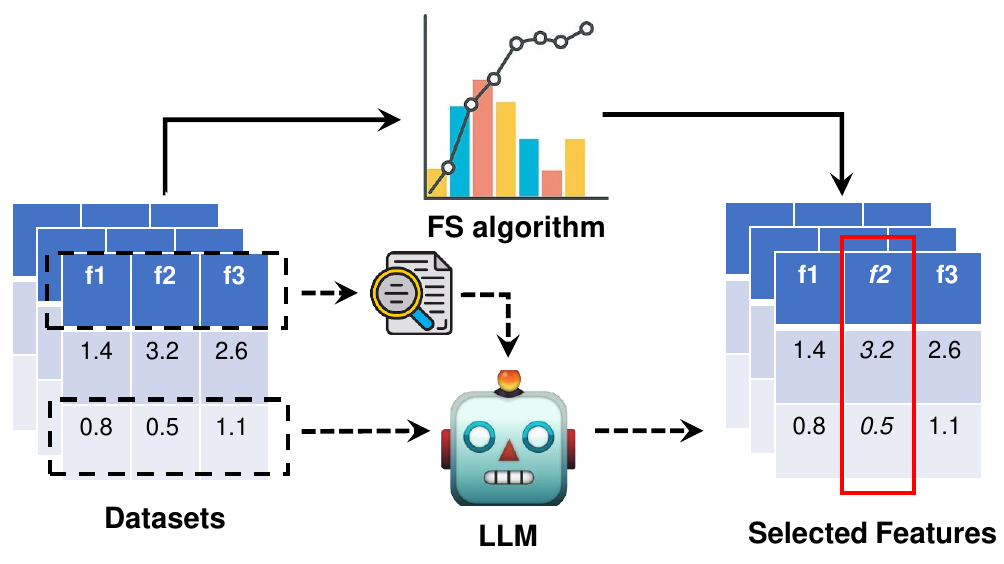}
    \caption{Comparison of traditional feature selection (FS) algorithms and LLM-based methods. Instead of requiring the whole dataset to make statistic inference, recent works prompt LLMs to select features in an efficient way. This is often achieved in a (\textit{\textbf{i}}) data-driven, or (\textit{\textbf{ii}}) text-based way.}
    \label{fig:overview}
\end{figure}

In this work, our objective is to thoroughly explore and understand LLMs-based feature selection methods from a data-centric perspective.
The conclusions and insights drawn from this exploration can provide insightful guidance for real-world applications where different types of resources and data are available.
To begin with, we categorize the prompting strategies in previous studies~\cite{choi2022lmpriors, jeong2024llm,liu2024ice,hanlarge} for LLMs-based feature selection into two groups: (\textit{\textbf{i}}) data-driven methods, which provide specific samples to LLMs~\cite{liu2024ice, hanlarge}, and (\textit{\textbf{ii}})~text-based methods, which incorporate feature and task descriptions into the instruction~\cite{choi2022lmpriors, jeong2024llm}.
These two prompting strategies require different data types: data-driven methods rely on sample points from datasets to do statistical inference while text-based methods need descriptive context for better semantic association between features and target variables. Figure~\ref{fig:overview} presents an overall comparison between the abovementioned methods and traditional feature selection algorithms.
These differences make us curious about how LLMs perform with each of them under different data availability settings.

We conduct extensive experiments to explore the two methods in both classification and regression tasks with different LLMs in various sizes (E.g. GPT-4, ChatGPT and LLaMA-2). 
\underline{\textbf{\textit{A key finding}}} based on the results is that, text-based feature selection using LLMs is more effective and stable across various low-resource settings. Additionally, it shows a more pronounced scaling law with respect to the size of LLMs compared to data-driven approaches.
Furthermore, we carried out a comparative evaluation between text-based feature selection using LLMs and traditional feature selection methods. \underline{\textbf{\textit{A general observation}}} is that, the text-based approach is relatively more robust and competitive across different resource availability settings.

Based on the abovementioned findings, we further explore the \textit{applicability} of text-based feature selection with LLMs in a medical application.
Specifically, we focus on the prediction of survival time for cancer patients~\cite{tomczak2015review,wissel2022survboard}, which is a crucial task to evaluate both patient health and treatment effectiveness.
To enhance the LLMs' understanding of medical-specific gene names, we developed a \underline{\textbf{R}}etrieval-\underline{\textbf{A}}ugmented \underline{\textbf{F}}eature \underline{\textbf{S}}election \textbf{(RAFS)} method that leverages descriptions from the National Institutes of Health (NIH) as auxiliary context.
Experiment results demonstrate our RAFS's effectiveness in performing effective feature selection while safeguarding patient's privacy.
Finally, we outline the existing challenges and potential opportunities in employing LLMs for feature selection.

To summarize, our contributions in this work are as follows:

\begin{itemize}
    \item We propose a general taxonomy for the existing LLMs-based feature selection methods, splitting them into data-driven and text-based methods.
    \item Through an analysis under varying data availability conditions, we identify the strengths and weaknesses of these two methods, finding that text-based approaches are more effective and robust.
    \item We showcase the utilization of the text-based feature selection method with LLMs in a real-world medical application and introduce RAFS, a method designed to handle domain-specific feature selection with LLMs.
    \item We systematically analyze the existing challenges and potential future directions for using LLMs in feature selection, providing further insights and guidelines for future studies.
\end{itemize}

\section{Related Work}
\vspace{0.35cm}

\subsection{Feature Selection}
\vspace{0.35cm}

Feature selection is the process of identifying and selecting the most relevant and important features or variables from a dataset to improve the performance and efficiency of a machine learning model~\cite{dash1997feature,guyon2003introduction,chandrashekar2014survey,li2017feature}. These feature selection methods can be generally categorized into three groups: filter, wrapper, and embedded approaches. Filter methods~\cite{lazar2012survey} first rank features by performing correlation analysis and then selecting the most important ones for the following learning step. Typical filter methods include mutual information~\cite{lewis1992feature,ding2005minimum}, Fisher score~\cite{hart2000pattern,gu2011generalized} and maximum mean discrepancy~\cite{song2012feature}. By contrast, wrapper methods~\cite{kohavi1997wrappers} use heuristic search strategies to identify a feature subset that optimally enhances the performance of certain prediction models (e.g., sequential selection~\cite{luo2014sequential} and recursive feature elimination ~\cite{guyon2002gene}). For embedded approaches, it works together with specific machine learning models in the training phase by adding various regularization items in the loss function to encourage feature sparsity~\cite{tibshirani1996regression,yuan2006model}.

\subsection{Feature Selection with LLMs}
\vspace{0.35cm}

There are already some works exploring the adaptation of LLMs in feature selection.~\cite{choi2022lmpriors} try to extract the relevant knowledge from LLMs as the task prior to performing feature selection, reinforcement learning and casual discovery. For feature selection, they design a prompt to instruct GPT-3~\cite{brown2020language} to generate whether given features are important by answering ``Yes" or ``No". Following them,~\cite{jeong2024llm} expand the LLMs-based feature selection and propose three different pipelines that directly utilize the generated text output. They also conduct extensive experiments in evaluation across various model scales and prompting strategies. Besides, some studies devise more complex pipelines with LLMs in feature selection and feature engineering.~\cite{liu2024ice} introduce an In-Context Evolutionary Search (ICE-SEARCH) in Medical Predictive Analytics (MPA) applications. It involves recurrently optimizing the selected features by prompting LLMs to perform feature filtering based on test scores.~\cite{hanlarge} employ LLMs as feature engineers to produce meta-features beyond the original features and combine them with simple machine learning models to improve predictions in downstream tasks. In this work, we aim to explore and understand LLMs in performing feature selection from a data perspective, offering further insights and hints for the adaptation of LLM-based feature selectors in real-world applications.

\begin{figure}[!t]
    \centering
    \includegraphics[width=8cm]{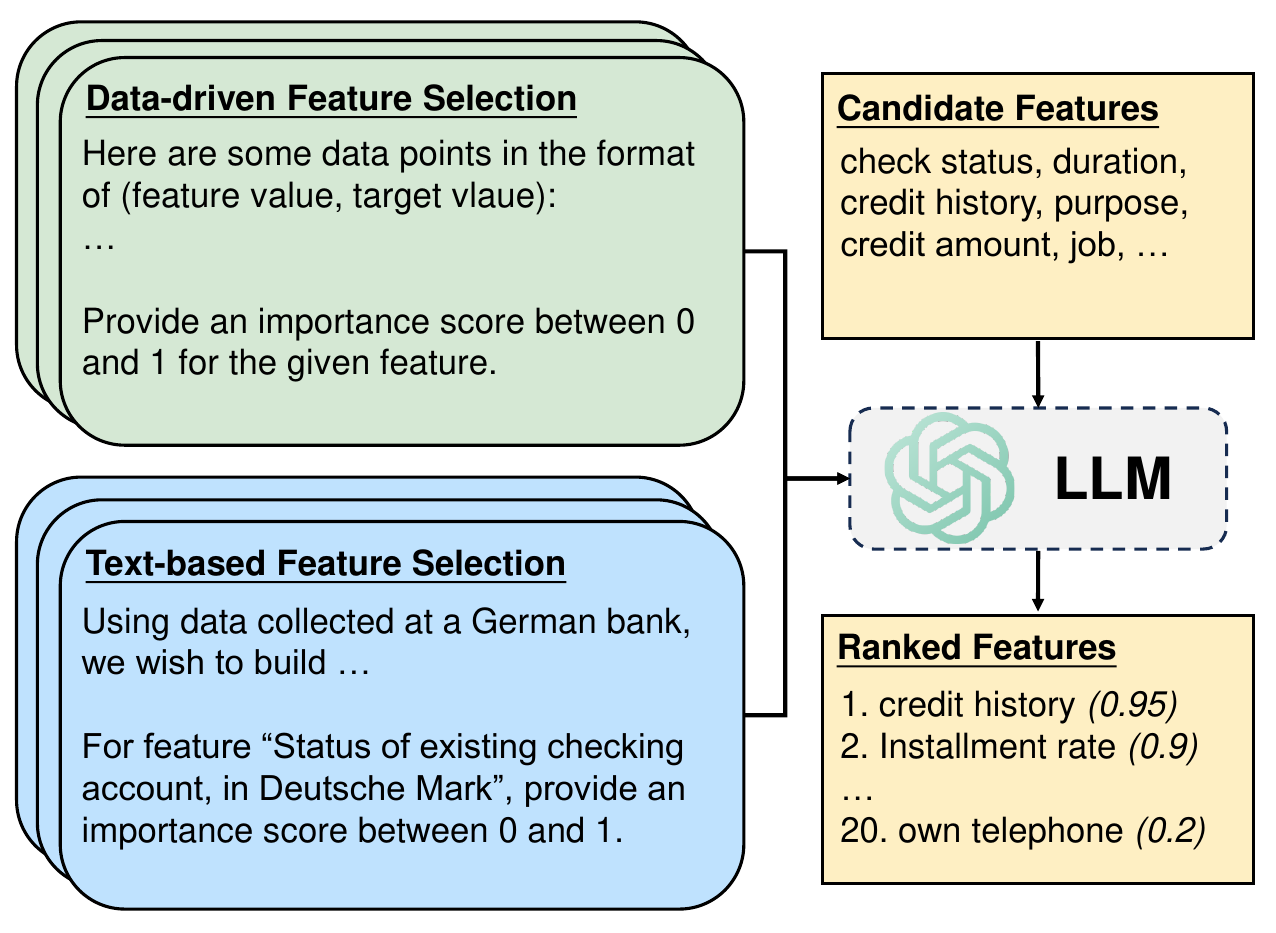}
    \caption{Prompting strategies for data-driven and text-based feature selection methods with LLMs.}
    \label{fig:prompt}
\end{figure}

\section{A Data-centric Taxonomy}
\vspace{0.35cm}

Given a pre-trained LLM $M$, we follow the scoring-based method proposed by~\cite{jeong2024llm}, which prompt $M$ to generate an importance score $s_i$ for the given feature/ concept $f_i$ in the dataset $d$:
\begin{equation}
    s_i = {\rm M}(P_{f_i}),\quad i\in \{1,...,l\},
\end{equation}
where $l$ is the total number of the features in dataset $d$. $P_{f_i}$ refers to the specific prompt we use to generate the importance score. We will discuss two methods for constructing prompts in Sections~\ref{Data-driven Feature Selector} and \ref{Text-based Feature Selector}, each focusing on different capabilities of LLMs.
Figure~\ref{fig:prompt} demonstrates the detailed prompting strategy for each of them.
\subsection{Data-driven Feature Selection}
\label{Data-driven Feature Selector}
\vspace{0.35cm}
Recently, LLMs have been employed to directly handle numeric data, demonstrating their capabilities in numerical prediction and analytics~\cite{gruver2024large,jin2023time}.
Therefore, we build a data-driven feature selection method with LLMs by providing both features' value $n_{f_i}$ and the value of the target variable $n_{y}$.
Intuitively, LLMs are supposed to infer the correlation and perform statistical analysis to determine the importance of the given feature in the dataset.

To be more specific, assume there are $m$ samples available in the dataset $d$, we first build the sample pairs $SP_i$ using values of the $i_{th}$ feature and target variable:
\begin{equation}
    SP_i = \{(n_{f_i}^j,n_{y}^j)\},\quad i\in \{1,...,l\}, j\in \{1,...,m\}.
\end{equation}
Then, we curate the prompt $P_{f_i}$ using $SP_i$ as few-shot examples and other instruction context $C$:
\begin{equation}
    P_{f_i}^{Data} = {\rm prompt}(C, SP_i),
\end{equation}
here $prompt$ is a function to concatenate the information and build a fluent instruction for LLMs.

\begin{figure*}[!t]
    \centering
    \includegraphics[width=18cm]{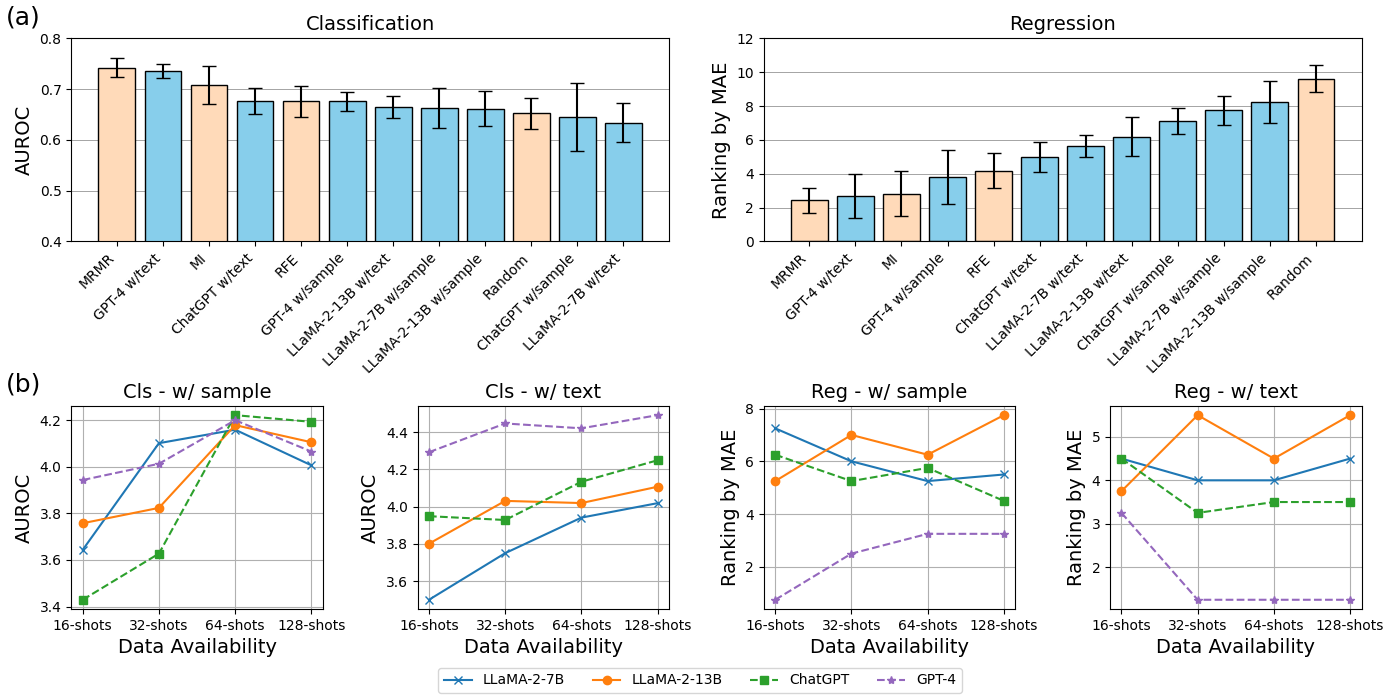}
    \caption{(a) Average AUROC (left; higher is better) and ranking by MAE (right; lower is better) across all datasets. (b) Each LLM's feature selection results, separated by task types (CLS and REG) and selection methods (w/sample and w/text).}
    \label{fig:result_by_method}
\end{figure*}

\subsection{Text-based Feature Selection}
\label{Text-based Feature Selector}
\vspace{0.35cm}

Another line of work~\cite{choi2022lmpriors, jeong2024llm} tries to employ the extensive semantics knowledge in LLMs~\cite{li2024contextualization} to perform feature selection.
Specifically, they incorporate detailed dataset descriptions in the prompt, instructing LLMs to semantically distinguish the importance of a given feature using their inherent knowledge and experience.

In our studies, we consider two concrete descriptive contexts: dataset description ($des_{d}$) and feature description ($des_{f_i}$). The dataset description includes the task's objective, details about the dataset's collection, and an explanation of the target variable. The feature description focuses on detailing and clarifying the feature to be scored.

Formally, we build prompts by integrating the abovementioned information:
\begin{equation}
    P_{f_i}^{Text} = {\rm prompt}(C, des_{d}, des_{f_i}).
\end{equation}

We give specific instruction examples for the two feature selection methods in Appendix~\ref{Detailed Instruction}.

\section{Analyses}
\vspace{0.35cm}

\subsection{Experiment Settings}
\vspace{0.35cm}

In this section, we evaluate the performance of the LLM-based feature selection methods using various datasets and models. 

\textbf{Models.} Below are the LLMs used in our experiment.

\begin{itemize}
    \item LLaMA-2~\cite{touvron2023llama}: 7B parameters.
    \item LLaMA-2~\cite{touvron2023llama}: 13B parameters.
    \item ChatGPT~\cite{OpenAI2023Introducing}: $\sim$175B parameters\footnote{$\sim$ denotes the estimated size~\cite{jeong2024llm} of closed-source LLMs}.
    \item GPT-4~\cite{achiam2023gpt}: $\sim$1.7T parameters\footnotemark[\value{footnote}].
\end{itemize}

We use the ``gpt-4-turbo-2024-04-09'' and ``gpt-3.5-turbo-0125 models via API calling. For LLaMA-2, we do local inference with the checkpoints available from Huggingface, namely ``llama-2-70b-chat-hf'' and ``llama-2-13b-chat-hf''.

\textbf{Compared Methods} As the main methods to be analyzed in this section, we use ``w/ data'' and ``w/ text'' to represent the data-driven and text-based feature selection methods. We also compare the LLM-based feature selection methods with the following traditional feature selection baselines:

\begin{itemize}
    \item Filtering by Mutual Information (MI)~\cite{lewis1992feature}.
    \item Recursive Feature Elimination (RFE)~\cite{guyon2002gene}.
    \item Minimum Redundancy Maximum Relevance selection (MRMR)~\cite{ding2005minimum}.
    \item Random feature selection.
\end{itemize}

\begin{table}[ht]\centering
\begin{tabular}{lcc}
\hline
Dataset     & \# of samples & \# of features \\ \hline
Adult       & 48842         & 14             \\
Bank        & 45211         & 16             \\
Communities & 1994          & 102            \\
Credit-g    & 1000          & 20             \\
Heart       & 918           & 11             \\
Myocardial  & 686           & 92             \\
Diabetes    & 442           & 20             \\
NBA         & 538           & 28             \\
Rideshare   & 5000          & 18             \\
Wine        & 6497          & 11             \\ \hline
\end{tabular}
\caption{Statistics of the datasets used.}
\label{tab:dataset}
\end{table}

\textbf{Datasets.} In our evaluation, we consider both classification and regression tasks. For the classification task, we use six datasets: Adult~\cite{asuncion2007uci}, Bank~\cite{moro2014data}, Communities~\cite{misc_communities_and_crime_183}, Credit-g~\cite{kadra2021well}, Heart\footnote{\href{https://kaggle.com/datasets/fedesoriano/heart-failure-prediction}{https://kaggle.com/datasets/fedesoriano/heart-failure-prediction}} and Myocardial~\cite{misc_myocardial_infarction_complications_579}. For the regression task, we use four datasets: Diabetes~\cite{efron2004least}, NBA\footnote{\href{https://www.kaggle.com/datasets/bryanchungweather/nba-player-stats-dataset-for-the-2023-2024}{https://www.kaggle.com/datasets/bryanchungweather/nba-player-stats-dataset-for-the-2023-2024}}, Rideshare\footnote{\href{https://www.kaggle.com/datasets/aaronweymouth/nyc-rideshare-raw-data}{https://www.kaggle.com/datasets/aaronweymouth/nyc-rideshare-raw-data}} and Wine~\cite{asuncion2007uci}. Detailed statistics of datasets are given in Table~\ref{tab:dataset}.

\textbf{Implementation Details.} For each dataset, we fix the feature selection ratio to 30\%. We vary the data availability for evaluations with 16-shot, 32-shot, 64-shot, and 128-shot configurations. The test performance is measured using a downstream L2-penalized logistic/ linear regression model, selected via grid search with 5-fold cross-validation. We use the area under the ROC curve (AUROC) to evaluate classification tasks and mean absolute error (MAE) for regression.

\subsection{Result Analysis}
\label{Result Analysis}
\vspace{0.35cm}

\begin{figure*}[!t]
    \centering
    \includegraphics[width=18cm]{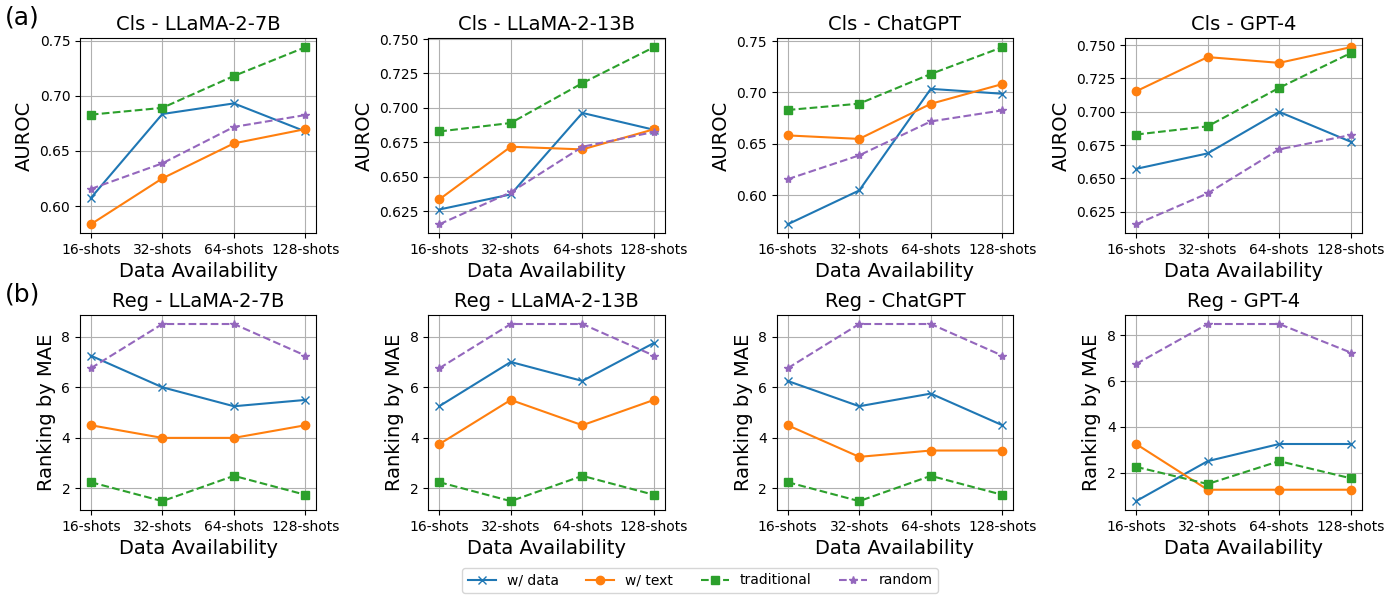}
    \caption{(a) Each feature selection method's results in the classification task, categorized by different LLMs; for each method, we add an error bar to represent its standard variant among various data availabilities. (b) Each feature selection method's results in the regression task, categorized by different LLMs. In each sub-figure, we include the average performance of traditional data-driven methods and the random selection method for comparison.}
    \label{fig:result_by_model}
    \vspace{-1mm}
\end{figure*}

We present our main experimental results in Figure~\ref{fig:result_by_method} and Figure~\ref{fig:result_by_model} for analyzing, and highlighting the following findings for answering the \textsc{Research Question}:

\textbf{Finding 1: Text-based feature selection is more effective than data-driven ones with LLMs in low-resource settings.} As results demonstrated in Figure~\ref{fig:result_by_method} (a), almost in every LLM and task (except LLaMA-2-7B in classification), the performance of small machine learning models with the text-based feature selection method surpasses that of the data-driven feature selection method. This finding is consistent when we delve into feature selection methods' performance in each data availability, as depicted in Figure~\ref{fig:result_by_model}. Additionally, in Figure~\ref{fig:result_by_method} (a), we notice for the same LLM, the text-based feature selection method usually leads to a smaller standard variant among various data availability settings. This further underscores the robustness and independence of the text-based feature selection method with respect to sample size.

\begin{table}[ht]
\centering
\begin{tabular}{lcc} \hline
             & AUROC          & Ranking by MAE \\ \hline
MI           & 0.779          & \textbf{1.75}  \\
RFE          & 0.758          & 3.50           \\
MRMR         & \textbf{0.798} & 2.25           \\ 
GPT-4 w/text & 0.783          & 2.50           \\ \hline
\end{tabular}
\caption{Feature selection results in the full dataset with traditional data-driven methods and ``GPT-4 w/text''.}
\label{fig:full shot}
\end{table}

\textbf{Finding 2: Text-based feature selection with the most advanced LLMs achieves comparable performance with traditional feature selection methods in every data availability setting.} In Figure~\ref{fig:result_by_method} (a), we observe that while GPT-4 with the text-based feature selection method performs slightly below the best traditional method (MRMR), it still demonstrates comparable performance, making it a competitive feature selection method in few-shot scenarios. However, when the LLM backbone is switched to less capable models, such as LLaMA, the text-based selection method shows a significant performance drop. Additionally, we experiment on the full dataset using `GPT-4 w/text' alongside three traditional feature selection methods, and found that GPT-4 with the text-based method remains competitive even in the full-shot scenario.

\textbf{Finding 3: Data-driven feature selection using LLMs struggles when number of samples increases.} An interesting phenomenon we observed is a significant performance drop in the classification task when the sample size increases from 64 to 128 using the data-driven feature selection method (Figure~\ref{fig:result_by_method} (b)). This drop is consistently observed across all four LLMs, indicating that each model generates poorer feature subsets as the sample size grows. We attribute this issue to LLMs struggling with processing long sequences, a challenge highlighted in many previous studies~\cite{dong2023bamboo,liu2024lost}. This limitation constrains the effectiveness of data-driven feature selection, which is why we did not include it in the full-shot experiment.

\textbf{Finding 4: Text-based feature selection exhibits a stronger scaling law with model size compared to data-driven feature selection with LLMs.} We investigated how scaling laws in model size affect feature selection capabilities. In Figure~\ref{fig:result_by_method} (b), we observe a clear correlation between the size of LLMs and their text-based feature selection capabilities. In contrast, while GPT-4 shows significantly superior performance in data-driven feature selection, the other three LLMs do not clearly follow the scaling law. This suggests that text-based feature selection is a reliable approach that can be enhanced by using powerful LLMs.

\section{Survival Prediction - A Case Study}
\vspace{0.35cm}

We use a biomedical task to showcase the utilization of LLMs-based feature selection in real-world applications.
Survival time prediction~\cite{tomczak2015review, wissel2022survboard} aims to predict cancer patients' survival time based on their physical and physiological indicators, playing a critical role in patient risk management and boosting treatment selection.
One of the significant challenges in survival prediction datasets is the huge volume of features (e.g., there are around 20,000 gene expression features in the TCGA~\cite{tomczak2015review} dataset).
While previous studies performed data-driven feature selection methods such as principal component analysis (PCA) to address this issue~\cite{wissel2023systematic}, as we mentioned in Section~\ref{Introduction}, It would cause serious privacy concerns for both patients and hospitals.

Impressed by the competitive performance and sample-free nature of text-based feature selection with LLMs, here we adopt it in the survival prediction application.
In our preliminary experiments, we found LLMs have difficulties in directly understanding the domain-specific feature name (e.g., gene ID).
Therefore, we borrow insights from retrieval-aug-mented generation (RAG) with LLMs~\cite{gao2023retrieval,chen2024benchmarking,li2024dalk} and propose \underline{\textbf{R}}etrieval-\underline{\textbf{A}}ugmented \underline{\textbf{F}}eature \underline{\textbf{S}}election \textbf{(RAFS)} to efficiently handle these biomedical-specific feature names.
Specifically, we retrieve meta information (e.g., official full name, summary and annotation information) about each feature name from the online National Center for Biotechnology Information (NCBI)\footnote{\href{https://www.ncbi.nlm.nih.gov/}{https://www.ncbi.nlm.nih.gov/}} and provide this information to LLMs as the support document for better feature selection.

\subsection{Experiment Settings}
\vspace{0.35cm}

We conduct experiments using the Lung Adenocarcinoma (LUAD) dataset in The Cancer Genome Atlas (TCGA) benchmark~\cite{tomczak2015review}.
Akin to~\cite{wissel2023systematic}, we use clinical indicators and gene expression as the full feature set and fix the feature selection ratio to be 30\%.
We use PriorityLasso~\cite{klau2018priority} as our machine learning backbone and report three metrics: Antolini’s Concordance (Antolini’s C)~\cite{tomczak2015review}, Integrated Brier score (IBS)~\cite{graf1999assessment} and D-Calibration (D-CAL)~\cite{haider2020effective}, all of which are commonly-used metrics for survival prediction.

\subsection{Result Analysis}
\vspace{0.35cm}

\begin{table}[ht]
\centering
\begin{tabular}{lccc}
\hline
              & Antolini’s C$\uparrow$    & IBS$\downarrow$             & D-CAL$\downarrow$           \\ \hline
PriorityLasso & 0.6306          & 0.1863          & 1.8518          \\
\quad w/ random     & 0.6516          & 0.1833          & 2.0255          \\
\quad w/ RAFS       & \textbf{0.6566} & \textbf{0.1830} & \textbf{1.7666} \\ \hline
\end{tabular}
\caption{Experiment results in TCGA-LUAD. We add random selection as the baseline to compare our RAFS with.}
\label{tab:survival prediction}
\end{table}

As the results show in Table~\ref{tab:survival prediction}, we find that even training the model on a randomly selected subset yields slightly better performance than training on the full feature set.
This implies the huge volume of features in TCGA-LUAD negatively impacts model performance, highlighting the importance of feature selection.
Moreover, we notice feature selection with our RAFS leads to significant performance improvements and consistently outperforms the random selection baseline. These findings suggest that RAFS is an effective approach for handling privacy-sensitive and large-scale biomedical datasets.

\section{Outlook}
\vspace{0.35cm}

In this section, we discuss potential opportunities for LLMs in feature selection, aiming to provide guidelines and hints for future works.

\textbf{Synergy of LLMs-based and traditional feature selection.} As we discuss in Sections~\ref{Introduction} and~\ref{Result Analysis}, text-based feature selection with LLMs is competitive and resource-efficient compared with traditional feature selection methods.
However, each approach relies on different sources of information—specific samples or context descriptions to perform feature selection. This diversity in information utilization makes them complementary. It would be valuable to explore how to combine text-based and traditional feature selection methods to create more effective and robust feature selection systems across various data availability scenarios. Also, it would be interesting to explore the synergy of text-based and data-driven methods to further enhance LLMs-based feature selection under resource constrains.

\textbf{Data-driven analysis with Agentic LLMs.} In Section~\ref{Result Analysis}, we conclude that poor statistical inference capabilities in long-sequence input hinder LLMs in data-driven feature selection.
While this finding implies the sole adaptation of LLMs may not be enough for performing data-driven feature selection, the introduction of agent-based LLMs should be considered as an alternative~\cite{xi2023rise,wang2024survey}.
These methods equip LLM with various tools~\cite{paranjape2023art,yang2024gpt4tools,schick2024toolformer} and APIs~\cite{patil2023gorilla,liu2024mmfakebench}, enabling them to execute actions and plans to solve complex and multi-step problems.
However, there are only a few works that focus on the development of agentic LLMs as data engineers and analytics~\cite{hong2024data,fang2024large,trirat2024automl}, for actively performing various features or data processing with the assistance of statistical tools or software.
Research in this direction will be valuable for enhancing and evaluating LLMs from analytical and statistical perspectives.

\textbf{Foundation models for feature/data engineering.} Many recent works have developed various foundation models in many data mining and machine learning fields, such as graph learning~\cite{liu2023towards,mao2024graph,xia2024opengraph} and time series prediction~\cite{rasul2023lag,jin2023time}.
A large foundation model for feature/ data engineering should be able to understand different types of information from the datasets and perform efficient manipulation and processing~\cite{cui2024tabular} to prepare appropriate data for downstream models/ applications.
Developing such a foundation model would greatly benefit the data mining and machine learning communities by providing a unified, easy-to-use interface for complex data processing tasks.

\section{Conclusion}
\vspace{0.35cm}

In this study, we explore feature selection methods based on LLMs from a data-centric perspective. We categorize existing LLM-based feature selection approaches into two main types: data-driven, which relies on statistical inference from specific samples, and text-based, which utilizes the extensive knowledge of LLMs for semantic association. Our experiments and analyses reveal that text-based feature selection with LLMs outperforms data-driven methods in terms of effectiveness, stability, and robustness. Based on these findings, we introduce a Retrieval-Augmented Feature Selection (RAFS) method designed to manage large volumes of domain-specific feature candidates in the context of cancer survival time prediction. Additionally, we provide a comprehensive analysis of the current challenges and potential opportunities at the intersection of LLMs and feature selection/engineering in Section 6, aiming to offer insights and guidance for future research in this area.

\section*{Acknowledgments}

The material in this presentation is based upon work supported by the U.S. Department of Homeland Security under Grant Award Number, 17STQAC00001-08-00, the U.S. Office of Naval Research (ONR) under grant N00014-21-1-4002, and the National Science Foundation (NSF) under grants IIS-2229461. The views and conclusions contained in this document are those of the authors and should not be interpreted as necessarily representing the official policies, either expressed or implied, of the U.S. Department of Homeland Security and the National Science Foundation.

%
\bibliographystyle{abbrv}
\bibliography{sigproc}  
%
%
\newpage
\onecolumn
\appendix
\section{Detailed Instruction}
\label{Detailed Instruction}

\begin{table*}[ht]
\centering
\begin{tabular}{c}
\hline
\begin{tabular}[c]{@{}l@{}}\color{red}/* Main System Prompt */ \\For the given feature, your task is to provide a feature importance score (between 0 and 1; larger value indicates greater\\ importance).\\ \\ \color{blue}/* Specific Sample Vlaues */ \\Here are some data points in the format of (feature value, target value), please refer to this to determine how informative\\ the feature is in predicting the target value:\\ (\textless{}0, no)\\ (no checking, no)\\ (\textless{}0, no)\\ (\textless{}0, no)\\ (0\textless{}=X\textless{}200, no)\\ (\textless{}0, no)\\ (\textgreater{}=200, no)\\ (\textless{}0, no)\\ (no checking, yes)\\ (no checking, yes)\\ (0\textless{}=X\textless{}200, yes)\\ (0\textless{}=X\textless{}200, yes)\\ (no checking, yes)\\ (0\textless{}=X\textless{}200, yes)\\ (0\textless{}=X\textless{}200, yes)\\ (\textless{}0, yes)\\\\ \color{red}/* Output Format Instruction */ \\ Here is an example:\\ ```\\ Question: What is the importance score for the given feature\\ Answer: The importance score is 0.9\\ ```\\ \\ \color{red}/* Main User Prompt*/ \\Question: What is the importance score for the given feature\\ Answer: The importance score is\end{tabular} \\
\hline
\end{tabular}
\caption{Detailed instruction for data-driven method in Credit-g dataset.}
\end{table*}

\begin{table*}[ht]
\centering
\begin{tabular}{c}
\hline
\begin{tabular}[c]{@{}l@{}}\color{blue}/* Dataset-specific Context */\\ Context: Using data collected at a German bank, we wish to build a machine learning model that can accurately predict \\ whether a client carries high or low credit risk (target variable). The dataset contains a total of 20 features (e.g., credit\\ history, savings account status). Prior to training the model, we first want to identify a subset of the 20 features that\\ are most important for reliable prediction of the target variable.\\ \\ \color{red}/* Main System Prompt */\\ For each feature input by the user, your task is to provide a feature importance score (between 0 and 1; larger value \\ indicates greater importance) for predicting whether an individual carries high credit risk and a reasoning behind how \\ the importance score was assigned.\\ \\ \color{red}/* Output Format Instructions */\\ The output should be formatted as a JSON instance that conforms to the JSON schema below.\\ \\ As an example, for the schema "properties": "foo": "title": "Foo", "description": "a list of strings", "type": "array", \\ "items": "type": "string", "required": {[}"foo"{]} the object "foo": {[}"bar", "baz"{]} is a well formatted instance of the schema.\\ The object "properties": "foo": {[}"bar", "baz"{]} is not well-formatted.\\ \\ Here is the output schema:\\ ```\\ \{"description": "Langchain Pydantic output parsing structure.", "properties": \{"reasoning": \{"title": "Reasoning", \\ "description": "Logical reasoning behind feature importance score", "type": "string"\}, "score": \{"title": "Score",\\ "description": "Feature importance score", "type": "number"\}\}, "required": {[}"score"{]}\}\\ ```\\ \color{blue}/* Demonstration */\\ Here is an example output:\\ -Variable: Installment rate in percentage of disposable income\\ \{"reasoning": "The installment rate as apercentage of disposable incomeprovides insight intoa person’s financial \\ responsibility and capability.This percentage can be seen as a measure of how much of a person’s available income \\ is committed to repaying their debts. If this rate is high, it might indicate that the person is taking more debt than \\ they can comfortably repay and may hint a talack off inancial responsibility, implyinghigher credit risk. If this rate\\ is low, it likely indicates that the person can manage their current financial obligations comfortably, implying \\ lowercredit risk. Thus, the score is 0.9.", "score": 0.9\}\\ \\ \color{red}/*Main User Prompt*/\\ Provide a score and reasoning for "Status of existing checking account, in Deutsche Mark." formatted according to the \\ output schema above:\end{tabular}\\
\hline
\end{tabular}
\caption{Detailed instruction for text-based method in Credit-g dataset.}
\end{table*}

\end{document}